\documentclass[]{ecai}  %

\usepackage{graphicx}
\usepackage{latexsym}
\usepackage{multicol}
\usepackage[linesnumbered,ruled]{algorithm2e}
\usepackage{comment}

\ecaisubmission      %

\paperid{1070}        %

\usepackage{tablefootnote}
\usepackage[linesnumbered,ruled]{algorithm2e}
\newcommand{\specialcell}[2][c]{%
  \begin{tabular}[#1]{@{}c@{}}#2\end{tabular}}

\begin{document}

\begin{frontmatter}

\title{Integrating Diverse Knowledge Sources for \\ Online One-shot Learning of Novel Tasks}

\author{\fnms{James R.}~\snm{Kirk}%
\thanks{Corresponding Author. Email: james.kirk@cic.iqmri.org}}
\author{\fnms{Robert E.}~\snm{Wray}%
}
\author{\fnms{Peter}~\snm{Lindes}} %
\author{\fnms{John E.}~\snm{Laird}%
}
\address{Center for Integrated Cognition, IQM Research Institute, Ann Arbor, MI USA}

\begin{abstract}
Autonomous agents are able to draw on a wide variety of potential sources of task knowledge; however current approaches invariably focus on only one or two. Here we investigate the challenges and impact of exploiting diverse knowledge sources to learn online, in one-shot, new tasks for a simulated office mobile robot. The resulting agent, developed in the Soar cognitive architecture, uses the following sources of domain and task knowledge: interaction with the environment, task execution and search knowledge, human natural language instruction, and responses retrieved from a large language model (GPT-3). We explore the distinct contributions of these knowledge sources and evaluate the performance of different combinations in terms of learning correct task knowledge and human workload. Results show that an agent's online integration of diverse knowledge sources improves one-shot task learning overall, reducing human feedback needed for rapid and reliable task learning.
\end{abstract}

\end{frontmatter}

\section{Introduction}
Typical AI systems are designed to exploit one or possibly two sources of knowledge, such as pre-programmed task knowledge (e.g., expert systems), experience in the world (e.g., reinforcement learning), processed human-language artifacts (e.g., large language models; LLMs), general planning capabilities, human-curated domain knowledge (e.g., ontologies), prior examples of task performance (e.g., imitation learning, deep learning), or direct interaction with humans (e.g., Interactive Task Learning; ITL). Each knowledge source presents different strengths and weaknesses. 

We hypothesize that an agent using multiple, diverse knowledge sources can exploit the strengths of individual sources while mitigating their weaknesses. To research this hypothesis, we explore Interactive Task Learning \cite{gluck_interactive_2019_custom,laird_interactive_2017,chai2018language,li2020interactive,kirk_learning_2019_custom,scheutz2018recursive}, where an embodied agent learns new tasks from human natural-language instruction, but where many other knowledge sources (such as those listed above) can contribute. In this paper, we specifically explore the challenges and contributions of diverse knowledge sources, including large language models. This approach presents three challenges. First, knowledge from different sources must be integrated during \textit{online} performance without pre-training. Second, task learning requires the learning and integration of \textit{many} types of task knowledge. Third, the learning must be in \textit{one shot,} so that instruction and training are not repeated.

\begin{figure}
    \centering
    \includegraphics[width=0.9\columnwidth]{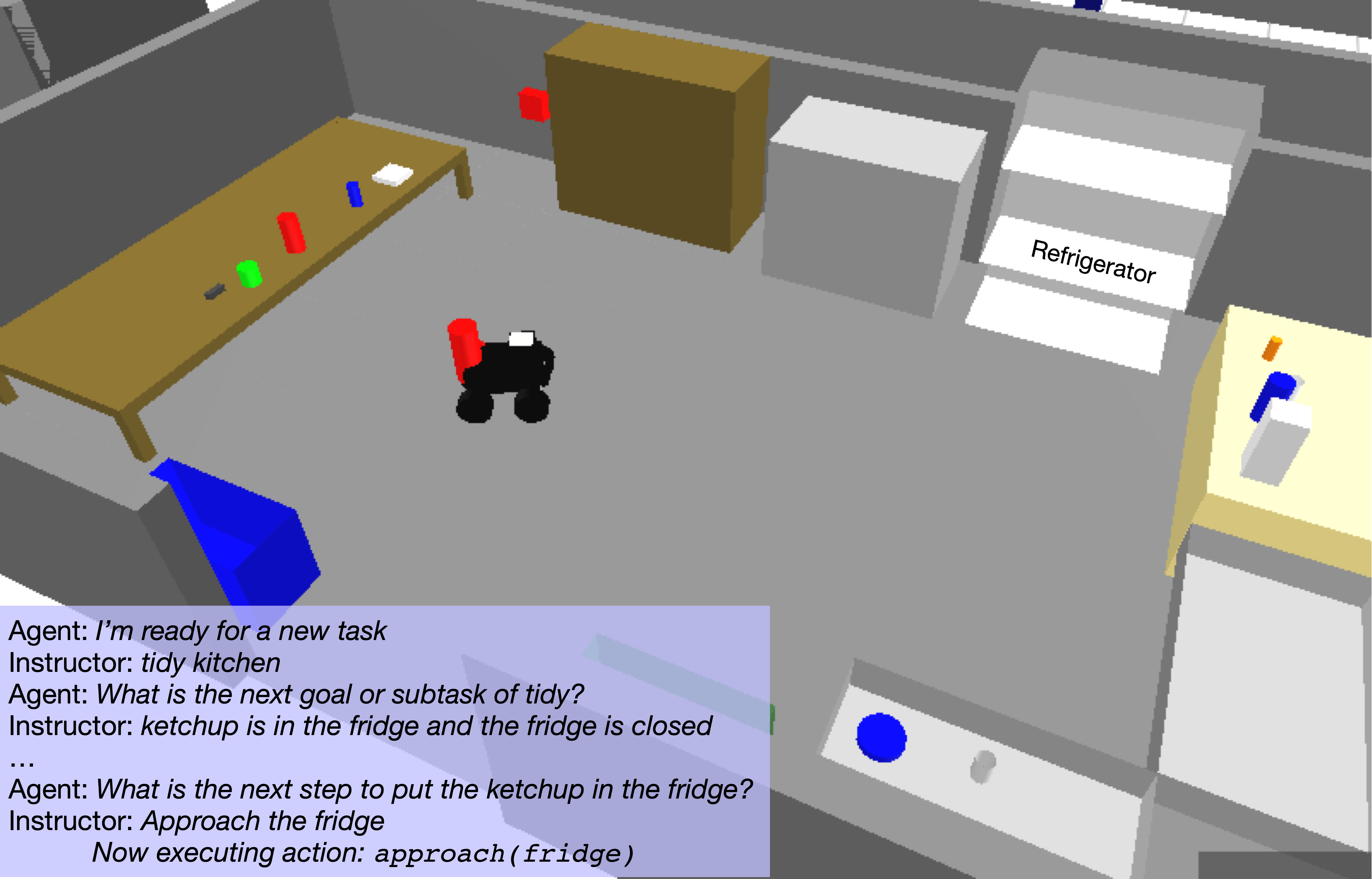}
    \caption{Example task: Learning to tidy a kitchen.}
    \label{fig:ketchup}
\end{figure}

We explore this new approach using an agent embodied as a mobile robot in a simulated, continuous office kitchen environment. Figure~\ref{fig:ketchup} illustrates a specific scenario, learning to tidy a kitchen, the learning task used throughout the remainder of the paper. Initially, the ITL agent uses a combination of its embodied perception and human instruction to learn to put the ketchup away (and then the remaining objects) in one shot. However, this approach requires that an instructor provide detailed and precise instructions, which can be cognitively demanding and time consuming. Internal search can ameliorate this issue (e.g., determining the steps needed to get the ketchup into the refrigerator), although it requires explicit goals and internal models of the available actions. 

In contrast, large language models (LLMs) ``contain'' vast knowledge, including the goals and actions associated with many tasks. Our ITL agent (described below) uses GPT-3 \cite{brown2020} to generate potential goal descriptions ({\tt ketchup in the fridge}) and actions ({\tt pick up ketchup}). Although LLM knowledge is broad and deep, it can also be inappropriate or incorrect for an agent's specific context. To mitigate inherent LLM unreliability, the agent deploys template-based prompting in combination with instructor feedback.

Below, we describe the online novel task learning problem, relevant prior work from the ITL and LLM research communities, and  potential benefits and costs of these knowledge sources for ITL. We then introduce our approach and outline how online interpretation of human instruction, search, and querying and interpreting responses from LLMs are integrated in the agent to learn to perform the target task in one shot. Experiments performed with different combinations of these knowledge sources demonstrate the positive impact of integration during task learning, qualitatively reducing the user input required, while still successfully learning novel tasks. 

\section{Novel Task Learning: Related Work}
There are many dimensions of variation in task learning, such as the types and variations of tasks learned, the types of task knowledge learned, the number of examples required to learn a task, the sources of available knowledge, and so on. Our focus is on online one-shot learning, where multiple, diverse tasks are learned through real-time interaction from multiple sources, including a LLM. Thus, exclusively experience-based methods, such as reinforcement learning (RL), which require extensive training, are not considered. 

Table \ref{tab:related_work} summarizes analysis of related work on ITL approaches, LLM approaches, and our approach. We analyze each approach in terms of the following dimensions: can it learn multiple tasks, from multiple sources, in one shot, online, from an LLM, and support knowledge transfer. For knowledge transfer, we contrast learning generalized task knowledge (applicable both in future identical scenarios and also in novel situations) to approaches that directly encode/memorize action sequences for a specific situation (or do not learn representations that can be applied in future contexts).
ITL is presented as a single category as most recent ITL work falls into the same categories for these dimensions. The LLM work that is closest to ours is Huang et al. \cite{huang2022inner_ijcai} which integrates human-in-the-loop learning with a LLM, but does not learn entirely online or in one shot.

The foundation for our approach is ITL
\cite{chai2018language,kirk_learning_2019_custom,mininger_expanding_2021,scheutz2018recursive}, which relies primarily on a human to teach an embodied agent in real-time through natural interactions, including language, gestures, demonstrations, and even sketching \cite{hinrichs2014x} (multi-source). The types of tasks vary from games and puzzles, to robot navigation and manipulation tasks (multi-task). 
ITL agents learn goal knowledge (final states to be achieved), control knowledge (when to select specific actions), grounding knowledge for making connections from internal concepts and actions to perception and motor control, and task decomposition knowledge. They encode the knowledge learned into internal representation, such that they can be used in the future (during another task or within the same task) to avoid relearning knowledge for the same or similar situations (knowledge transfer).
Many of these systems employ search/planning to supplement instruction. All learn new tasks from single examples (one shot) but none employ LLMs.

\begin{table}[t]
\centering
\begin{tabular}{|l|c|c|c|c|c|c|}
\hline
Approach    & 
     \rotatebox[origin=r]{90}{ Multi-task?} & 
     \rotatebox[origin=r]{90}{ Multi-source?} & 
     \rotatebox[origin=r]{90}{ One shot?} &
     \rotatebox[origin=r]{90}{ Online?} &
     \rotatebox[origin=r]{90}{ Uses LLM?} &
     \rotatebox[origin=r]{90}{ Know. Transfer?}
     \\
     \hline \hline
     Recent ITL  & Y & Y & Y & Y & N & Y     \\
     \hline
     SayCan \cite{ahn2022can_ijcai} & Y & Y & Y & N & Y & N   \\ %
     \hline
     Logeswaran et al. \cite{logeswaran_few-shot_2022_ijcai} & Y & Y & N & N & Y & N  \\ %
     \hline
     Huang et al. \cite{huang2022inner_ijcai} & Y & Y & N & Y & Y & N     \\ %
     \hline
     Gato \cite{reed_generalist_2022_custom} & Y & Y & N & N & Y & Y*    \\ 
     \hline
    This paper & Y\tablefootnote{Although our approach is multi-task, the experiment focuses on a single task with multiple subtasks.} & Y & Y & Y & Y & Y     \\
     \hline
    
\end{tabular}
\caption{Approaches to multi-task, multi-source task learning.}
\label{tab:related_work}
\end{table}

SayCan \cite{ahn2022can_ijcai} extracts knowledge from LLMs to learn many physically-grounded tasks (multi-task) in an embodied robot. It uses knowledge of its embodiment and perception to construct customized prompts, which is similar to our LLM prompting approach. Other than the LLM and embodiment, its other primary knowledge source is RL (multi-source), which is used to learn novel action primitives through pre-training (offline). Beyond this pre-training the learning from the LLM is done in one shot.  It does not learn or encode a representation of task knowledge to support task performance or learning in future scenarios (knowledge transfer).

Logeswaran et al. \cite{logeswaran_few-shot_2022_ijcai} investigate using LLMs to predict plans for multiple tasks in real-world environments, using environment context (multi-source) to rank LLM predictions and improve the relevance of retrievals.  Unlike our approach, they perform few-shot learning, and require training data for learning subgoals and the ranking model rather than learning entirely online. They learn an agent policy, but do not demonstrate that knowledge is encoded in such a way that it can be transferred to future task performance scenarios.

Huang et al. \cite{huang2022inner_ijcai} explore using LLMs for planning of multiple tasks in embodied robotic environments, but not task learning. They investigate injecting feedback from the environment and a human (multi-source) into LLM prompts to improve instruction completion. Like us, they focus on integrating human, LLM, and search knowledge. Unlike our work, they emphasize the dynamic nature of the interaction, including allowing the human to interrupt LLM-based planning.
Performance on ``unseen'' tasks ranges from very low to 76\%. Although they rely on some pre-training for robot skills, their main contribution relevant to this work is online. Similar to the baseline work they build on, SayCan (as above), they do not learn a representation of task knowledge that supports knowledge transfer.

Gato \cite{reed_generalist_2022_custom} uses a LLM to learn many different types of tasks (multi-task), including Atari games and robot manipulation tasks. It is trained on many different sources of different modalities (multi-source), in the form of datasets targeted for the different tasks it learns. Gato is a few-shot learner that was trained offline, but demonstrates learning on a wide variety of tasks. Models learned are trained over many tasks. Gato does directly address and demonstrate some generalization and transfer between tasks. However, transfer sometimes has a negative impact and improving transfer is a direction of future research.

\begin{table}[t]
\centering
\begin{tabular}{|l|c|c|c|c|c|c|}
\hline
     Knowledge Source & 
     \rotatebox[origin=r]{90}{ Goal K} & 
     \rotatebox[origin=r]{90}{ Control K} & 
     \rotatebox[origin=r]{90}{ Relevance} &
     \rotatebox[origin=r]{90}{ Breadth} & 
     \rotatebox[origin=r]{90}{ Correctness} & 
     \rotatebox[origin=r]{90}{ Affordability} 
     \\
     \hline \hline
     Human Instruction  & + & + &  + & + & + &  -   \\
     \hline
     Search & - & + &  + & - & + &  +  \\
     \hline
     LLM & + & + & ? &  + & ? & +   \\
     \hline
\end{tabular}
\caption{Assessment of knowledge sources for task learning}
\label{tab:knowledge_dimensions}
\end{table}

\section{Assessment of Knowledge Sources}
Here, we analyze the three knowledge sources explored in this paper (human instruction, search, and LLMs), theorizing qualitative costs and benefits for task learning. This analysis, summarized in Table \ref{tab:knowledge_dimensions},  provides a basis for subsequent experiments. The columns list two types of knowledge to be learned and four properties of knowledge sources:
\begin{itemize}
    \item Goal K: Does the source provide knowledge for defining goals, such as the many elements of ``tidy kitchen?'' 
    \item Control K: Does the source provide knowledge that informs what actions should be used when?
    \item Relevance: How directly relevant is the knowledge to the specific tasks the agent must learn and the specific situations in which it is learning those tasks? 
    \item Breadth: What is the extent of the available knowledge? 
    \item Task Correctness: Can the agent perform the task correctly after learning? Correct performance is critical in real-world task environments. 
    \item Affordability: What are the external costs (human time \& effort) for using the knowledge source? A solution that requires extensive human input will not scale.

\end{itemize}
Cells contain our initial assessment of the knowledge source properties: strength (`+'), weakness (`-'), or ambiguity (`?'). 

\begin{table*}[t]
    \centering
    \begin{tabular}{p{0.1\textwidth} p{0.85\textwidth}}
    \hline 
    \emph{Agent-created Prompt}     &   (EXAMPLES)(TASK) Task name: deliver package. Task context: I am in mailroom. Aware of package addressed to Gary, package is in mailroom. \par
    (RESULT)The goal is that the package is in Gary's office(END RESULT) \par
Steps: \par
1. Pick up package addressed to Gary \par
2. Go to Gary's office \par
3. Put package onto desk in Gary's office \par 
\par (END TASK)(END EXAMPLES) \par
(TASK) Task name: tidy kitchen. Task context: I am in kitchen. Aware of plastic bottle on counter.
\\[0.25ex] \cline{2-2} \emph{GPT-3} \par \emph{Response}     &  (RESULT)The goal is that the plastic bottle is in the recycling bin.(END RESULT)  Steps: \par
1. Pick up plastic bottle from counter \par
2. Put plastic bottle in recycling bin(END TASK) \\
\hline
    \end{tabular}
    \caption{Example of an instantiated prompt with demarcation  and the resulting response from GPT-3 (Temperature=0).}
    \label{tab:gpt_Example}
\end{table*}

\subsection{Human Instruction}
Via instruction, as in ITL, a human can teach an agent a new task by specifying intermediate goals and subgoals or by leading the agent through the task step by step. A human can also interactively define other concepts such as novel objects, relations, and properties; and suggest actions to take in current or hypothetical situations. The strength of instruction is that a human often has complete, correct, and contextually-relevant knowledge, including sharing the same environment and having awareness of the task and the agent's embodiment. The most significant limitation is the requirement that there is a knowledgeable and attentive human.

\subsection{Search}
When a task goal is known, an agent can often use search to achieve it. Search results in task-relevant, actionable control knowledge that is grounded in the agent's embodiment and environment. It relies on task-independent but embodiment-specific action models, which can restrict its breadth of application. Its applicability is also only tractable when the number of available actions is limited and solutions do not require deep searches. The knowledge it produces is correct (assuming correct action models).

\subsection{Large Language Models}
Breadth of knowledge is a significant strength of LLMs. A goal of this effort is to determine whether an agent can use the task context to construct an LLM query so that resulting responses are sufficiently relevant and correct to be useful for task learning. In the familiar situation of an office kitchen, we expect relevant responses. However, because LLMs generate responses based on statistical patterns of words in the training set, there is no guarantee that responses will be relevant or correct. Individual queries, parameterized for the agent's situation, are computationally and monetarily inexpensive. However, queries that need to be repeated many times for complex task learning can be monetarily costly, and it can be computationally expensive to test and verify responses. We seek approaches for integrating LLMs that ensure correctness, relevance, and affordability.

\section{Approach}
Our approach builds on an existing one-shot, ITL learning system which learns over 60 games and puzzles, indoor navigation tasks such as interior guard, and associated concepts to ground new terminology \cite{kirk_learning_2019_custom,mininger_expanding_2021}. Learning new tasks expands agent capabilities, and knowledge learned from previous tasks, such as common subgoals or shared task concepts, transfer to new tasks. 
The agent is implemented in the Soar cognitive architecture \cite{laird_soar_2012_custom} and has been embodied in four physical robots and many simulation environments \cite{mohan_learning_2014,kirk_learning_2019_custom,mininger_expanding_2021}.

After receiving a new task, the agent prompts the instructor to describe goals required to execute the task.
The agent parses and grounds natural-language instructions to create an internal interpretation and asks clarifying questions when necessary. From these internal representations of goals, subtasks, and supporting concepts, the agent then attempts to perform the task and achieve the goals, using search if needed, and learns both goal knowledge (semantic structures) and control knowledge (policy rules) for the task.
Once the agent has learned a new task, it can immediately perform it.

\subsection{Human Instruction} 
In the agent described above, the human initially drives the interaction, naming a new task for the agent to perform. The agent then requests a description of the goal (or a procedure for non-goal-oriented tasks), and will request the definition of an unknown word or a known word that the agent cannot ground to the current situation. To learn a task, such as ``tidy kitchen,'' the instructor describes the primary task (tidy), followed by instructions to repeat subtasks until all objects have been cleared from each location.

\subsection{Search} 
For learning novel tasks, it is difficult to have task-independent heuristics or cost-functions beyond depth of search. The ITL agent uses iterative deepening so that simple solutions are found quickly and the total search time is limited by a fixed threshold. A side-effect of search in Soar is that control knowledge is learned via Soar's chunking mechanism \cite{laird_soar_2012_custom}.
That is, for future, similar tasks, the agent has knowledge that specifies what actions should occur when, obviating the need for repeated search in comparable situations.

\subsection{Large Language Model}
In this research, we use GPT-3 \cite{brown2020} for a combination of its wide use, breadth of knowledge, accessibility, and relative affordability.\footnote{Total cost of GPT-3 for experiments in this paper was $<$ \$10 USD.} We have investigated GPT-4, but are not using this model as it does not currently support the logprobs feature (which provides relative token probabilities important to goal and action retrieval, as discussed below). 
We have extended the ITL agent so that it can construct a query (prompt) to GPT-3 customized for its situation and task-learning goal.

An example prompt is shown in Table \ref{tab:gpt_Example}. Everything above the intervening horizontal line is the prompt constructed by the agent; everything below is the response from GPT-3. The prompt, including parenthetic delimiters such as ``(EXAMPLES),'' is presented as (just) text. Even though they are not special commands, they lead GPT-3 to produce structured responses, such as the one shown in the example.

We use template-based prompting strategies \cite{olmo_gpt3--plan_2021,kirk2022improving,10.1145/3411763.3451760} to support learning task goals and actions. The initial part of the prompt template provides an example task. The example specifies a task to learn (deliver package), the task context (in a mailroom), and an object to execute the task on (package addressed to Gary). This task description is followed by the response desired from the LLM, which is the task goal denoted with (RESULT) and then steps to achieve that goal. The final line of the prompt is instantiated from the template, using the task, current location context, and object description: 
\begin{quote}
(TASK) Task name: \textbf{?task}. Task context: \\ \textbf{?current-location}. Aware of \textbf{?object-description}. 
\end{quote}
The prompt's syntax and word usage reflects the agent's language capabilities; this structure biases GPT-3 to provide an interpretable, relevant response. It is also biased toward responses that include actions the agent knows (more below). The agent interprets LLM responses using the same language parsing and grounding process it uses in interactions with humans.

\section{Combining Knowledge Sources}
As indicated in Table \ref{tab:knowledge_dimensions}, none of the knowledge sources provides all the desired features. Human instruction comes close, but it requires human time and effort, which can be unaffordable. Thus, our goal is to explore how these sources can be combined to mitigate their limitations and capitalize on their strengths. In this section, we discuss integrations of pairs of sources as well the integration of all three knowledge sources. For each of these conditions, the human provides the name of the task and subtasks to learn. The procedure the agent uses for learning from the multiple sources is defined in Algorithm~\ref{alg:main} (conditional statements are used to vary different knowledge sources conditions). 

In all combinations, the human initiates learning by introducing (naming) the task and describing subtasks to achieve, e.g., clearing objects off the table (line~\ref{line:gettask}). For each object the agent senses in its current location (e.g., on the table), it records the relevant context (the task, object description, and location) before attempting to learn a goal for it when the goal is not known (lines 3-6). It does not need to relearn goals for another instance of an object of the same type (e.g., two plates on the table).
We now outline how the agent learns goal concepts and associated control knowledge for objects in each of these knowledge combinations.

\subsection{Search + Human Instruction}
The original agent used instruction to acquire descriptions of the goal (line 11), and search to find a sequence of actions to execute to achieve the goal (lines 16-20). When the goal is achieved, control knowledge for executing the task is learned. The agent learns generalized policy knowledge, enabling transfer to similar tasks. For example, after learning to put a plate on the table into the dishwasher and close it, when it encounters a metal-fork on the table and learns the same goal, the control knowledge learned for the plate applies and it does not need to perform search or ask for action steps.\footnote{See section 3 of Appendix for details on the learned representations}
If the search fails (i.e., hits the depth threshold without finding a solution), as in the following example for learning to clear objects from the table, the agent asks the human for the next step (line 26).\footnote{All examples in this section are from actual interactions between a human instructor and the agent.}
\begin{quote}
\textbf{Instructor}: If the object is a plastic bottle then the goal is that the object is in the bin.

\textbf{Agent}: What do I do next for clear?

\textbf{Instructor}: Pick up the plastic bottle.
\end{quote}
After the agent executes the step, if the goal is still not satisfied (line 15), search is invoked again (lines 16-20) and the process continues until the goal is achieved (bottle in bin).
This combination makes a meaningful difference when interacting with the agent. Search is sufficient for solving many tasks (given a goal) and thus it eliminates much of the tedium of instructing the agent (e.g., giving it instructions to pick up a bottle). However, it still requires instructions to describe the goal and to provide actions when search fails.

\subsection{Human Instruction + Large Language Model}
The agent constructs a prompt for both a goal and for actions, as was shown in Table \ref{tab:gpt_Example}.
The agent first retrieves potential goals from the LLM (line 8) by constructing a prompt using the context, as descrided above.\footnote{In the experiments, we used two prompt examples,``deliver package'' (as in the table) and a similar example about storing a package in a closet.}

The agent attempts to retrieve three unique goal responses (line 8), one with temperature 0 and two with higher temperatures (0.9, 1.0). In GPT-3, temperature controls sampling temperature, with temperature 0 producing near deterministic results. LLM responses are often identical: the agent will query up to 10 times (escalating to the highest temperature for the last 5 attempts) to try to retrieve three unique responses.
The agent then finalizes the choice (line 9), in this condition, by asking the instructor if the goal is correct, selecting the goal with highest average log probability.

\begin{quote}
\textbf{Agent}: For a glass in the dish rack is the goal that the glass is in the dishwasher and the dishwasher is turned on?

\textbf{Instructor}: No.

\textbf{Agent}: For a glass in the dish rack is the goal that the glass is in the cupboard and the cupboard is closed?

\textbf{Instructor}: Yes.
\end{quote}
If the human responds negatively, the agent retrieves the next highest probability response from the LLM and queries the human again. If the human responds in the affirmative, the agent transforms the goal from the LLM into a conditional goal statement for the object (e.g. ``If the object is a glass then the goal is that the object is in the cupboard and the cupboard is closed.'').  It then attempts to learn the goal from this natural language description using its existing language and goal learning abilities.

Once the goal is learned, the agent needs to learn the actions to achieve the goal. When search is not available (line 18), the agent retrieves action steps from the LLM (lines 22-23). The agent appends the goal response to the prompt in order to generate steps for that goal. Unlike goals, action retrieval is accomplished in two steps. In the first step, five single words are retrieved. The agent filters these responses based on their log probability. This step enables the agent to bias action responses toward actions (verbs) known to the agent. In the second step, each chosen word is then appended to the original prompt and a complete action statement is generated, as for goals\footnote{This function and others are described with detailed examples in Section 1 of the Appendix}. After retrieval, the agent again finalizes the choice (line 24) by selecting the most probable action and asking the instructor for confirmation.
If the responses retrieved from the LLM are exhausted, the agent requests the human give a description: 
\begin{quote}
\textbf{Agent}:  Should I ``Pick up plate from dish rack?''

\textbf{Instructor}: No.

\textbf{Agent}:  What do I do next for unload?

\textbf{Instructor}: Open cupboard.
\end{quote}
Ideally, the human provides only confirmation of the goal and task steps, decreasing human effort. However, there is no guarantee that the LLM responses will be correct, relevant, or grounded to the task, environment, and the agent's embodiment. In those cases the human needs to be correct errors and fill in missing steps by providing goal and action descriptions.

\begin{algorithm}[ht]
\SetKwProg{Fn}{Function}{:}{}
\SetKwFunction{GetTask}{get\_novel\_task\_from\_user}
\SetKwFunction{FindGoals}{retrieve\_goals}
\SetKwFunction{FinalizeGoal}{finalize\_choice}
\SetKwFunction{GetGoal}{get\_goal\_from\_user}
\SetKwFunction{IDSearch}{IDSearch}
\SetKwFunction{Execute}{execute\_actions}
\SetKwFunction{IterativeActionRetrieval}{retrieve\_actions}
\SetKwFunction{ConfirmAction}{confirm\_with\_user}
\SetKwFunction{GetAction}{get\_action\_from\_user}

\Fn{LearnNewTask()}{
    task $\leftarrow$ \GetTask{}\;
    \label{line:gettask}
    \While{some objects not evaluated}{
        choose a sensed object in current location\;
        context $\leftarrow$ task + object + location\; \label{line:getcontext}
        \If{no goal recognized for the chosen object}{
            \uIf{LLM condition}{
                pot\_goals $\leftarrow$ \FindGoals{context}\;
                goal $\leftarrow$ \FinalizeGoal{pot\_goals}\;
            }
            \Else{
                goal $\leftarrow$ \GetGoal{context}\; \label{line:getusergoal}
            }
        }
        \If{goal}{
            \While{goal not satisfied}{
                \If{action\_list nil and search condition}{
                      action\_list $\leftarrow$ \IDSearch{goal, depth}\;
                }
                \uIf{action(s) on action\_list}{
                    \Execute{action\_list}\;
                }
                \Else{
                    \uIf{LLM condition}{
                        pot\_acts $\leftarrow$ \IterativeActionRetrieval{goal, context}\;
                        action $\leftarrow$ \FinalizeGoal{pot\_acts}\;
                    }
                    \Else{
                        action $\leftarrow$ \GetAction{context}\;
                    }
                    action\_list $\leftarrow$ action\;
                }
            }
        }
    }
}
\caption{Procedure for learning a new task.}
\label{alg:main}
\end{algorithm}

\subsection{Search + Large Language Models}
When search is combined with the LLM, the agent queries the LLM only for subtask goals and uses search to find a solution, using the prompting process described above but only for subtask goals. The human only describes the initial task (e.g. ``tidy kitchen'') and subtasks (e.g., clear all objects off the table). 
In this combination, to finalize the goal (line 9), the most probable response for a goal from the LLM is used by the agent. If the goal retrieved from the LLM is correct, the agent finds a solution. If the goal is not correct but is achievable (in the current situation), the agent finds a solution, but the correct goal for the subtask is not achieved. If the goal retrieved is uninterpretable by the agent or is unachievable (e.g. closing a recycling bin that has no lid), then the agent, for experimentation purposes, records the failure and moves on to the next object.

In some cases, this combination will result in success without any human instruction beyond naming the task (e.g., tidy) and subtasks (e.g., clear). However, whenever the LLM falls short, a partial task failure (on the object) will occur. It is difficult to predict \textit{a priori} the frequency of these failures. We explore this question experimentally, as summarized in the subsequent section.

\subsection{Combined}
When all knowledge sources are available, the approach is similar to the Instruction and LLM combination, but the agent uses search to avoid asking for actions when possible.

The combination of all methods should result in more correct outcomes than the search and LLM condition and requires substantially less human intervention than the other combinations that use human instruction. Assuming this general analysis is correct, the most interesting question is the extent to which human instruction can be reduced.  If the combination of LLM, search, and instruction substantially reduces needs for human instruction, this result would be a compelling case for incorporating LLMs in interactive task learning (and other agent learning problems).

\section{Experiment Design and Results}
In this section, we describe the design and results of an experiment that evaluates the performance and costs of various combinations of the knowledge sources for online, one-shot learning of the ``tidy kitchen'' task.

\subsection{Experiment Design}
The task environment is a simulated office kitchen and mobile robot created in the APRIL MAGIC robot simulator. The mobile robot has a single arm and manipulator and can grasp and manipulate all objects relevant to the task to be learned (i.e., the task is achievable with its primitive capabilities).

Experiments use both the robot simulation and ``mental simulation.''  Mental simulation uses knowledge encoded in Soar to simulate the environment state and its dynamics. Mental simulation simplifies system and asset engineering (e.g., defining new objects, placing objects, etc.) and experiments are thus faster to create and run. Regardless of the environment used, the agent learns exactly the same task knowledge. Data presented below was generated using ``mental simulation.'' However, we also verified the learning in the robot simulation. The agent, using knowledge learned from mental simulation, successfully performs the tidy task when embodied in the simulated robot environment.
\begin{figure}[tb]
    \centering
    \includegraphics[width=0.90\columnwidth]{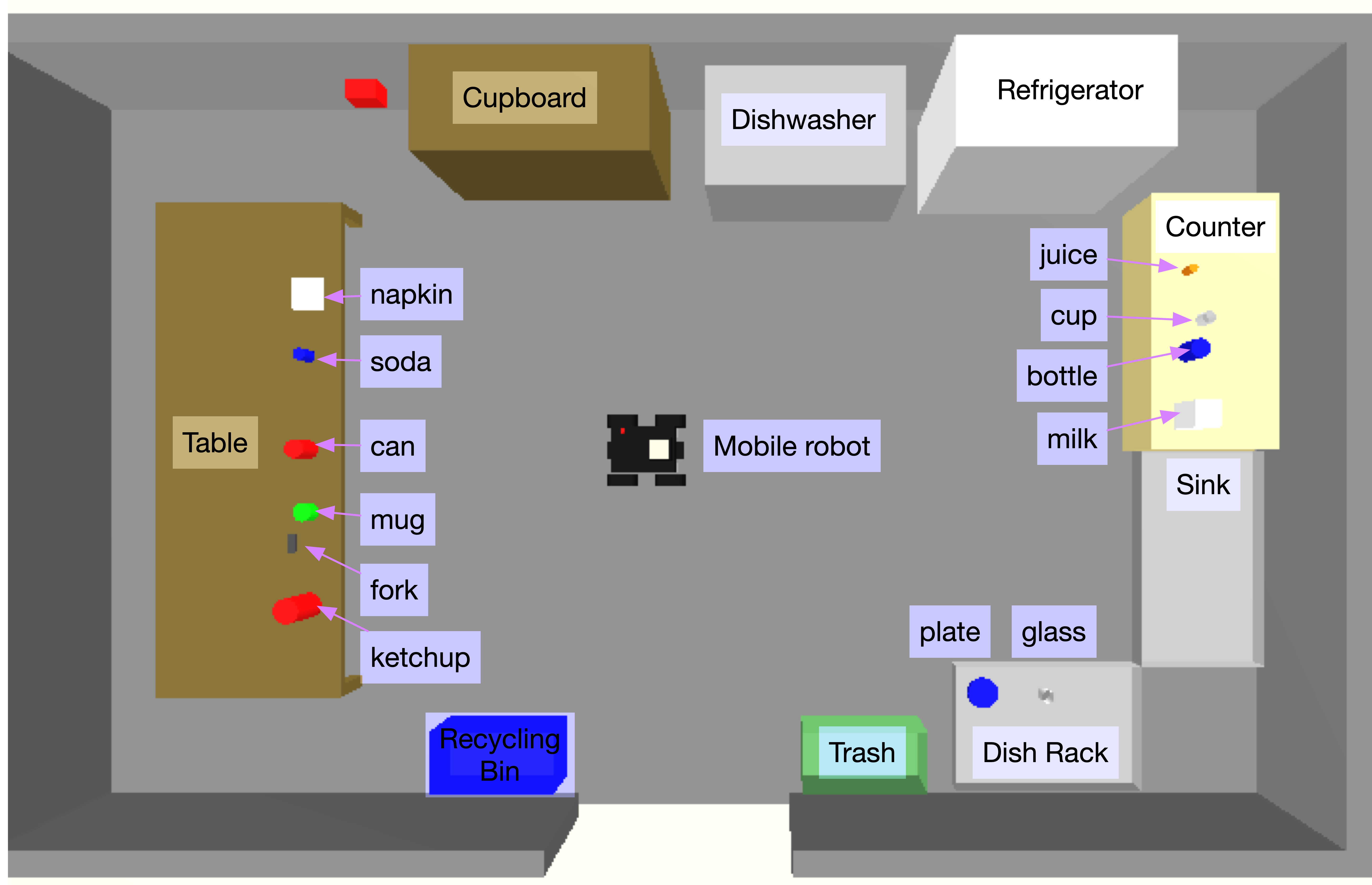}
    \caption{Example situation for ``tidy kitchen'' learning task.}
    \label{fig:initial_conditions}
\end{figure}
\subsubsection{Learning Task} 
The experimental task is to learn to tidy a kitchen in an office building. We chose this task because it is commonly used for robotic task learning, it offers a realistic use case (regardless of its \textit{a priori} capabilities, an office robot will need to be customized for a specific office building and the people who work in it), and its general familiarity makes it easy to convey the specific goals and actions being learned. 

In all the conditions, an experimenter (``instructor'') initiates the task learning by telling the agent it will learn to tidy a kitchen. 
The human then introduces subtasks, which include clearing, storing, and unloading all the objects. 
How the agent learns the goals and actions for each object differs between conditions (described in the next subsection).
Figure~\ref{fig:initial_conditions} is a representative task-learning scenario (with fewer objects and destinations for ease of illustration). 

Thirty objects are located in various places in the room (e.g., a paper plate on a table). The overall goal of ``tidy kitchen'' is to move each object to an appropriate location. Each type of object has a target location, determined by the type of object, its location, and its properties (e.g., the mug on the table goes into the dishwasher or sink but the glass in the dish rack should be placed in the cupboard, the agent cannot perceive directly whether the dishes are clean or dirty). The agent can perceive object types and properties, but the goal location for each object is unknown to the agent and the agent does not know that objects have goal locations until it is instructed.

Some goals require that the single-armed robot perform extra steps because it cannot open doors while holding an item. We expect that LLM responses will sometimes omit such steps (or mis-order them for the robot) %
because the LLM's knowledge will usually assume an embodiment and affordances that match those of human actors rather than those of a mobile robot \cite{ahn2022can_ijcai}. Our prompting strategy addresses this issue by including a sequence of actions in the prompt example (see Table \ref{tab:gpt_Example}) that are particular to the agent's embodiment. Such a prompt biases responses towards instructions to open things first. Thus, the agent can potentially constrain and influence LLM responses to reflect its specific embodiment.

In addition to relocating objects, five of the locations in the environment have doors or drawers that must be closed for successful task completion. For instance, the agent must learn to put the ketchup in the refrigerator and to close its door. 
Thus, there are 35 goal assertions to be achieved for ``tidy kitchen'' (30 objects to be moved + 5 objects that must be closed).
In sum, the experimental goal is for the agent to learn an executable representation of the tidy-kitchen task that allows the agent to perform the task in the same (or similar) environment in the future without further instruction.

\subsubsection{Experimental Conditions}
The experimental conditions are based on knowledge source combinations outlined in the previous section. For all experiments, we use GPT-3 text-davinci-002 with TopP set to 1 and Temperature varying between 0 and 1.0 as described above.
\begin{itemize}
\item Human Instruction: The human provides both goal descriptions (``milk in fridge'') as well as control knowledge in the form of action imperatives (``pick up milk'').
\item Instruction + Search: The agent receives goal descriptions from instruction and uses search to find a solution without any action instruction.
\item Search + LLM: The agent creates prompts to elicit goal descriptions from the LLM and uses search (max. depth of 4) to achieve them without instruction. The highest probability LLM responses are selected. We expect that this condition will result in failures when  LLM responses are inconsistent with the environment, agent's embodiment, and/or existing knowledge.
\item Instruction + LLM: The agent constructs prompts to the LLM for both goals and actions. The human answers yes/no questions to accept/reject the response. If 3 suggestions are rejected, the agent asks for a goal description or next action. 
\item Instruction + Search(max. depth=2) + LLM:  The LLM is prompted for both goals and actions. Human instruction is again used to accept/reject responses and provide descriptions after three LLM-generated suggestions are rejected. Search is used to find the goal state, but is deliberately limited to a depth of 2, so that when the limit is reached, the LLM is queried for actions.
\item Instruction + Search(max. depth=4) + LLM: Same as the previous condition except that the search depth is 4, which is sufficient to find a solution with a correct goal formulation for all objects.
\end{itemize}

\begin{table}[tb]
    \centering
    \begin{tabular}{|l|r|r|r|r|r|}
    \hline 
     Condition & \rotatebox[origin=c]{90}{\specialcell{Completion \\ Rate (\%)}} & 
     \rotatebox[origin=c]{90}{\specialcell{~Relevant LLM \\ ~Responses (\%)}} &
     \rotatebox[origin=c]{90}{~\# of Instruct.} & 
     \rotatebox[origin=c]{90}{\# of Words} & 
     \rotatebox[origin=c]{90}{\# Yes/No Inst.} \\ \hline\hline

Instruction & 100.0 &  -- &  76 & 757 &   0 \\ \hline
Inst+Search$_{d4}$ & 100.0 &  -- &  40 & 603 &   0 \\ \hline
Search$_{d4}$+LLM & 54.3 & 53.3 &  14 &  76 &   0 \\ \hline
Instruction+LLM & 100.0 & 59.1 & 112 & 348 &  88 \\ \hline
Inst+Search$_{d2}$+LLM & 100.0 & 40.3 &  94 & 384 &  67 \\ \hline
Inst+Search$_{d4}$+LLM & 100.0 & 30.2 &  70 & 360 &  43 \\ \hline
    
\hline
    \end{tabular}
    \caption{Summary of experimental results.}
    \label{tab:overall_results}
\end{table}

\subsubsection{Measures} The measures for evaluating the knowledge source properties (other than breadth) from Table~\ref{tab:knowledge_dimensions} are listed below. We expect outcomes to be consistent with that table but also to quantify relative costs and benefits for this task.

\begin{itemize}
    \item Correctness: Task completion rate (the number of goal assertions achieved / total number of goal assertions (35, as above)) is used as a proxy for correctness.
    \item Relevance: For these experiments, instruction and search (as we have defined them) are always relevant. We measure relevance in LLM conditions as the percentage of accepting (``yes'') responses given the total number of yes/no confirmation queries.
    \item Affordability: Human time is the largest factor impacting affordability. We assess affordability by number of instructions, number of words (because some instructions, such as goals, are more complex than others), and  Yes/No responses (used in Inst.+LLM conditions and should be easier than generating goals or actions).
  
\end{itemize}

\subsection{Result Discussion}
\begin{figure}[tb]
    \centering
    \includegraphics[width=0.98\columnwidth]{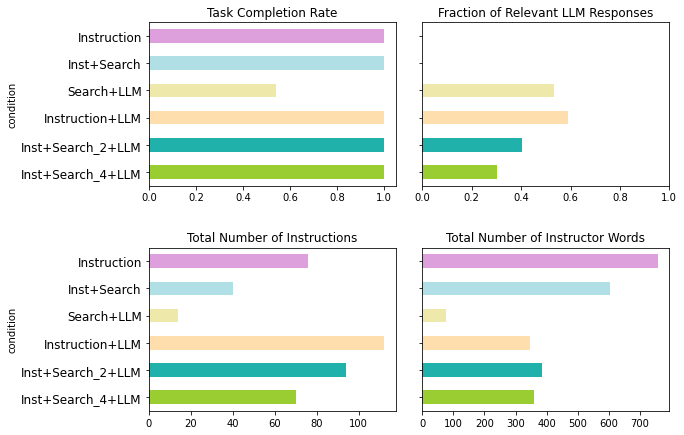}
    \caption{Summary of key measures for all experimental conditions.}
    \label{fig:overall_results_fig}
\end{figure}

Results for all conditions are provided in Table~\ref{tab:overall_results} and illustrated in Figure~\ref{fig:overall_results_fig}. We discuss each measure individually.

\subsubsection{Completion Rate} In all conditions other than Search+LLM, the agent fully learns all subtasks in one shot. When the agent is presented a similar scenario, it completes ``tidy kitchen'' without additional search, instruction, or LLM prompting. 
In the Search+LLM condition, 54\% of subtasks are completed. Failures arose when the LLM generated either an incorrect response (put a dirty mug in cupboard) or steps that could not be executed (``close bin'' when the bin lacks a lid). 

\subsubsection{Relevant Responses} 30-40\% of the responses generated by the LLM were accepted in Inst+LLM conditions. The relevancy rate in Inst+Search$_{d4}$+LLM is notably lower (30\%) than in the other LLM conditions. The LLM is significantly better at producing relevant actions (96\% correct for first try) than goals. For depth=4, no actions are generated, so relevancy is determined completely by the relevance of goal responses.

A similar dynamic explains comparatively high relevance in Search+LLM. Only one response from the LLM is retrieved for each goal (i.e., in contrast to up to 3 times in Inst+LLM conditions). This results in greater relevancy because the LLM tends to generate similar responses goals, often resulting in the rejection of multiple responses in the Inst+LLM conditions. Where the first response was used, of the 16 objects that were not successfully tidied, 3 were due to the LLM generating goals to close a bin without a lid (missing context), 3 were due to putting empty bottles and cans in the fridge (no awareness they were empty), and 2 were due to putting objects in the cupboard instead of the pantry.

\subsubsection{Affordability Measures} 
Because the LLM is used to supplement instruction, the mixed results in the number of instructions may at first seem inconsistent with expectations. However, the total number of instructor words decreases substantially in all cases of LLM use (roughly 50\%). Reductions are achieved because the LLM produces candidate goals and actions, which the instructor accepts or rejects via simple yes/no responses (rightmost column of Table~\ref{tab:overall_results}). A significant reduction in number of instructions occurs for Search+LLM because LLM-produced goals are accepted automatically (i.e., no human review); this condition also results in failures for many objects. While Inst+Search$_{d_4}$+LLM results in only 6 fewer total instructions compared to the baseline Instruction condition (76 vs. 70), it results in only 37 goal and action instructions (70 instructions - 43 yes/no), thus reducing the number of complex instructions by about half.

\section{Discussion and Conclusions}
Experimental results are consistent with the analysis and experimental hypothesis summarized in Table~\ref{tab:knowledge_dimensions}. 
The most important outcome is that even with unexceptional acceptance rates for goal responses from the LLM, when coupled with simple yes/no feedback from the human instructor and search (that mitigated the need for action instructions), the combined approach enabled 100\% task completion while substantially reducing human interaction. Further, by incorporating new goal, action selection, and control knowledge within the task-learning agent, the agent not only can perform the same tasks on re-presentation (one-shot learning) but can also transfer aspects of this learning to new objects and tasks.

LLMs are a promising tool for task-knowledge acquisition. 
However, they are not yet reliable enough to be used by themselves for task learning. 
Agent use of multiple knowledge sources enables learning task knowledge that is relevant, correct, and actionable for a specific situation that a robot or other agent may encounter. 
Furthermore, LLMs can significantly reduce the human interaction costs associated with interactive task learning.  

For future work, we plan to explore more diverse domains and complex tasks that involve more interactions between objects, rather than tasks where each object can be handled independently. Interactions will likely lead to a computational explosion in search costs (negligible in this study), furthering the need for other knowledge sources like an LLM. Taking inspiration from Huang et al. \cite{huang2022inner_ijcai}, we also plan to inject context for failures into prompts. The goal would be to include failures from previous queries in the prompt, to encourage greater variability in repeated queries for a given goal. We also plan to explore utilizing other sources of knowledge, including knowledge bases such as WordNet \cite{fellbaum_wordnet_1998}  and ConceptNet \cite{speer_conceptnet_2017}. We have performed an initial integration with WordNet to identify synonyms of words, so that a greater fraction of LLM responses can be interpreted by the agent.

\section*{Ethical Statement}

This work uses large language models which can present ethical and social risks \cite{weidinger_ethical_2021_custom} such as discrimination, exclusion, and toxicity or malicious uses. We consider each of these risks.

Because large language models are generative, depending on their corpus and training, they can produce language that reflects cultural biases, offensive stereotypes, derogatory usages, etc. In this work, where LLM queries are focused solely on producing goals and actions for a particular task environment, we have not seen \textit{any} responses from GPT-3 that include such language. 

In terms of exclusion, the specific task we have chosen does reflect (and mirror) a cultural specificity to US/Western settings, in that the items in the kitchen (and the labels used to describe them) are both specific to the English language and typical of the objects one would find in a kitchen in an office building. One of the long-term potential outcomes of this work (and ITL more generally) is that the robot is taught by human users in a particular setting, allowing the human to customize behavior to their specific setting (including its cultural context). Investigating whether this potential can be realized in the subject of future work.

Malicious use is also a potential risk, in that this research aims to enable human users to instruct robots to do their bidding. Users could theoretically instruct robots to cause direct harm to others, violate laws, etc. At this point in our research, this risk is minimal because implementations are confined to controlled, laboratory experiments. We are actively investigating in other work how an ITL agent can be both instructed while also following and conforming to both codified rules (like laws) and social norms to further mitigate the potential for malicious use.

\ack This work was supported by the Office of Naval Research, contract N00014-21-1-2369. The views and conclusions contained in this document are those of the authors and should not be interpreted as representing the official policies, either expressed or implied, of the Department of Defense or Office of Naval Research. The U.S. Government is authorized to reproduce and distribute reprints for Government purposes notwithstanding any copyright notation hereon. We would also like to thank contributors to the development of the ITL agent: Mazin Assanie, Elizabeth Goeddel, Aaron Mininger, Shiwali Mohan, and Preeti Ramaraj. 

\bibliography{ecai,zotero-transoar,bob-bibliography,ijcai23,aaai23}

\onecolumn

\appendix
\newcommand{\apptitle}[1]{\noindent{\centering\LARGE #1}}
\apptitle{\hspace{.35\linewidth} Technical Appendix}

\section{Pseudo-code for Individual Algorithms}
This section provides pseudo-code outlines of the primary components of the task learning process described in the paper. The tables provide schematic representations of the algorithms in functional form. The actual implementations of the algorithms here are realized within Soar. The LLM Query functions to retrieve LLM knowledge are invoked by the agent and are implemented in python using standard LLM APIs.

Soar provides a parallel knowledge retrieval process from both procedural and declarative memories and a decision process that selects operators and problem spaces based on the agent's context. The execution of a problem space in Soar is roughly comparable to the execution of a function in the below. However, the implementation does not require (or assume) the linear steps outlined here. For example, if an agent already has received a task and goal, it would skip immediately to determining actions to satisfy the goal (i.e., line 15 of Algorithm~\ref{alg:main}).

The tables summarize the agent's task-learning process, which includes prompt generation, response evaluation, and execution of chosen responses. The various experimental conditions are referenced in the functions as well. For example, in non-LLM experimental conditions, the agent asks the user for a goal and actions for each object in the task environment.

\begin{algorithm}[H]
\SetKwProg{Fn}{Function}{:}{}
\SetKwFunction{GetTask}{get\_novel\_task\_from\_user}
\SetKwFunction{FindGoals}{retrieve\_potential\_goals}
\SetKwFunction{FinalizeGoal}{finalize\_choice}
\SetKwFunction{GetGoal}{get\_goal\_from\_user}
\SetKwFunction{IDSearch}{IDSearch}
\SetKwFunction{Execute}{execute\_actions}
\SetKwFunction{IterativeActionRetrieval}{iterative\_action\_retrieval}
\SetKwFunction{ConfirmAction}{confirm\_with\_user}
\SetKwFunction{GetAction}{get\_action\_from\_user}

\Fn{LearnNewTask()}{
    task $\leftarrow$ \GetTask{}\;
    \While{some objects not evaluated}{
        choose a sensed object in current location\;
        \If{no goal recognized for the chosen object}{
            \If{LLM condition}{
                pot\_goals $\leftarrow$ \FindGoals{object, task, location}\;
                \tcp{Ask user, choose "best" option if no user}
                goal $\leftarrow$ \FinalizeGoal{pot\_goals}\;
            }
            \Else{
                goal $\leftarrow$ \GetGoal{object, task, location}\;
            }
        }
        \If{goal}{
            \While{goal not satisfied}{
                \tcp{Skip search if actions on queue or no search}
                \If{action\_list nil and search condition}{
                      action\_list $\leftarrow$ \IDSearch{goal, object, location, depth}\;
                }
                \If{action(s) on action\_list}{
                    \Execute{action\_list}\;
                }
                \Else{
                    \If{LLM condition}{
                        pot\_next\_actions $\leftarrow$ \IterativeActionRetrieval{goal, object, task, location}\;
                        action $\leftarrow$ \FinalizeGoal{pot\_next\_actions}\;
                        action\_list $\leftarrow$ action
                    }
                    \Else{
                        action $\leftarrow$ \GetAction{object, task, location, state}\;
                        action\_list $\leftarrow$ action
                    }
                }
            }
        }
    }
}
LearnNewTask()\;
\caption{Procedure for learning a new task.}
\label{alg:main2}
\end{algorithm}

There are five algorithms summarized here:
\begin{itemize}
    \item Algorithm~\ref{alg:main2}. The primary task-learning function. The agent receives a new task (``tidy kitchen'') and subtasks (e.g.,``clear objects on the table'') and then evaluates objects in its local environment to determine if those objects are relevant to the task. (This version is slightly expanded over Algorithm~\ref{alg:main} in the main body of the paper.)
    \item Algorithm~\ref{alg:goals}. This function retrieves potential goals from the LLM. It attempts to retrieve a prescribed number of unique goals from the LLM. 
    \item Algorithm~\ref{alg:actions}. This function is comparable to the goal retrieval function but instead retrieves a list of possible next actions.\footnote{Multiple actions are typically required to satisfy any goal but this function attempts to retrieve the ``next'' action, not the set of all actions.} The action-retrieval process first retrieves individual words and then biases selection of complete action statements towards action words that the agent recognizes.
    \item Algorithm~\ref{alg:unique_res}. This function attempts to retrieve unique responses from the LLM. It is used for retrieval of goals and actions. At low temperatures (less variability), the instantiated prompt tends to result in identical responses from the LLM. This function increases the temperature for queries when additional unique responses are needed.
    \item Algorithm~\ref{alg:finchoice}. Having retrieved responses from the LLM, the agent must choose one of the responses to execute. This function summarizes how the choice is made. In conditions where the user is present, the user is presented each generated response (from highest to lowest log probability as computed by the LLM) and asked a yes/no question to accept or to reject the response. If all retrieved responses are rejected, the user is asked to enter the goal or next action (using natural language text inputs). When no user is present, the agent uses the response with the greatest log probability.
\end{itemize}

\begin{algorithm}[hbt]
\SetKwProg{Fn}{Function}{:}{}
\SetKwFunction{ChooseTemplate}{choose\_template}
\SetKwFunction{Instantiate}{instantiate}
\SetKwFunction{GetUniqueResponses}{get\_unique\_responses}

\Fn{retrieve\_potential\_goals(object, task, location)}{
   \BlankLine
   \tcp{Instantiate relevant template}
   template\_type $\leftarrow$ goal\;
   template $\leftarrow$ \ChooseTemplate{template\_type}\;
   prompt $\leftarrow$ \Instantiate{template, object, task, location}\;
   
   \BlankLine
   \tcp{Algorithm parameters}
   \tcp{Desired number of unique responses}
   num\_responses $\leftarrow$ 3\;
   \tcp{Max number of tries}
   max\_attempts $\leftarrow$ 10\;
   \BlankLine
   return \GetUniqueResponses{prompt, num\_responses, max\_attempts}\;
}
\caption{Query LLM for potential goals for object and task.}
\label{alg:goals}
\end{algorithm}

\begin{algorithm}[H]
\SetKwProg{Fn}{Function}{:}{}
\SetKwFunction{ChooseTemplate}{choose\_template}
\SetKwFunction{Instantiate}{instantiate}
\SetKwFunction{QueryLLM}{query\_LLM}
\SetKwFunction{QueryLLMToken}{query\_LLM\_for\_single\_token}
\SetKwFunction{GetUniqueResponses}{get\_unique\_responses}

\Fn{iterative\_action\_retrieval(goal, object, task, location)}{
   \BlankLine
   \tcp{Instantiate relevant template}
   template\_type $\leftarrow$ action\;
   template $\leftarrow$ \ChooseTemplate{template\_type}\;
   prompt $\leftarrow$ \Instantiate{template, goal, object, task, location}\;
   
   \BlankLine
   \tcp{Algorithm parameters}
   low\_temp $\leftarrow$ 0.0\;
   known\_word\_threshold $\leftarrow$ 0.09\;
   unknown\_word\_threshold $\leftarrow$ 0.5\;
   num\_words $\leftarrow$ 5\;
   num\_actions $\leftarrow$ 2\;
   max\_attempts $\leftarrow$ 5\;
   
   \BlankLine
   word\_list $\leftarrow$ \QueryLLMToken{prompt, low\_temp, num\_words}\;
   
   \BlankLine
   \tcp{Remove words from list that are below threshold}
   \For{word in word\_list}{
        \If{word is known}{
            threshold $\leftarrow$ known\_word\_threshold\;
        }
        \Else{
            threshold $\leftarrow$ unknown\_word\_threshold\;
        }
        \If{logprob(word) $\geq$ threshold}{
            prompt $\leftarrow$ prompt + word\;
            pot\_actions $\leftarrow$ \GetUniqueResponses{prompt, num\_actions, max\_attempts}\;
                }
    }
    \KwRet{pot\_actions}\;
}
\caption{Query LLM for potential actions for object and task.}
\label{alg:actions}
\end{algorithm}
\begin{algorithm}[hbt]
\SetKwProg{Fn}{Function}{:}{}
\SetKwFunction{QueryLLM}{query\_LLM}

\Fn{get\_unique\_responses(prompt, desired\_number, max\_attempts)}{
   \BlankLine
   \tcp{Algorithm parameters}
   low\_temp $\leftarrow$ 0.0\;
   high\_temp $\leftarrow$ 0.9\;
   \If{desired response type is goal}{max\_temp $\leftarrow$ 1.0\;}
   \Else{
   {\tcp{Dont escalate temp for actions}
   max\_temp $\leftarrow$ 0.9\;}}

   \BlankLine
   \tcp{Get low-temp response}
   responses $\leftarrow$ \QueryLLM{prompt, low\_temp}\;
   tries $\leftarrow$ 1\;

   \tcp{Get responses at higher temp(s), add to list if unique}
   \While{len(responses) $<$ desired\_number and tries $\leq$ max\_attempts}{
        \If{tries $>$ max\_attempts / 2}{
            high\_temp $\leftarrow$ max\_temp\;
        }
        response $\leftarrow$ \QueryLLM{prompt, high\_temp}\;
        tries $\mathrel{+}= 1$\;
        \If{response not in responses}{
            responses $\leftarrow$ responses + response\;
        }
    }
    \KwRet{responses}\;
}
\caption{Attempt to retrieve a given number of unique responses to the prompt from the LLM}
\label{alg:unique_res}
\end{algorithm}

\begin{algorithm}[hbt]
\SetAlgoLined
\SetKwProg{Fn}{Function}{:}{}
\Fn{finalize\_choice(responses, type)}{
    \SetKwData{finalchoice}{final\_choice}
    \tcp{Order the responses by descending log probability}
    responses = sort(responses, type=descending, value=logprob)\;
    
    \BlankLine
    \If{not a user condition}{
        \tcp{If no user, return highest logprob response}
        \Return responses.top()\;
    }
    \Else{
        \tcp{Ask user to confirm, from most to least likely}
        \For{response in responses}{
            feedback = ask\_user\_for\_confirmation(response, type)\;
            \If{feedback is yes}{
                \Return response\;
            }
        }
        
        \BlankLine
        \tcp{No responses accepted. User to input.}
        \If{type is goal}{
            \Return get\_goal\_from\_user(object, task, location)\;
        }
        \Else{
            \Return get\_action\_from\_user(object, task, location)\;
        }
    }
}
\caption{Choose one response from the candidates}
\label{alg:finchoice}
\end{algorithm}

\section{Step-by-Step Example of Learning Process}
In this section, we summarize the agent learning process step-by-step, complete with all prompts and responses, for learning the goals and actions for an object observed by the agent, a ceramic-plate on the table.
This example is created under the experimental condition that uses all knowledge sources: Instruction+Search$_2$+LLM. With a search depth of 2, it is necessary to elicit some of the actions from the LLM as well as goals. In this condition, the human is asked to confirm goals and actions and, if all options are rejected, to provide a description of the goal or action (See Algorithm~\ref{alg:finchoice} and the example below).

\subsection{Learning a goal}
During learning to clear all the objects that are on the table for \texttt{tidy kitchen}, the agent attends to tidy a ceramic plate that is on the table (i.e., in this example, line 4 of Algorithm \ref{alg:main} results in a ceramic plate being the object of interest).\footnote{In the embodiment in this agent, the reasoning component of the agent directly receives \texttt{ceramic-plate} from its perceptual system. The agent treats \texttt{ceramic-plate} as an atomic object type. When the agent constructs prompts to the LLM, it sends ``ceramic-plate''.} 
The agent attempts to retrieve a goal to achieve for the ceramic-plate for this task (line 5). When it fails to retrieve such goal knowledge (in this LLM condition; line 6), the agent attempts to retrieve potential goals from the LLM (Algorithm~\ref{alg:goals}). 

The retrieval process first constructs a prompt (lines 2-4) and then sets the conditions for retrieving a number of unique responses from the LLM (lines 5-7).
Prompt construction draws on a library of prompt templates that support various agent task-learning needs. In this case, the agent is attempting to determine what it should do about the plate, so it chooses a template focused on eliciting goals from the language model (line 2). 

The template embeds example prompts and responses from other task domains that influence the form of the goal statement that the LLM produces. The examples bias the LLM to produce goal statements that have the same form as the examples do, which is important for the agent's ability to parse and to interpret the responses produced by the LLM. The templates the agent uses in this paper were developed through an empirical process but in a different task domain (an office environment). Further, the templates have been effective in a number of different domains \cite{wray_language_2021_custom,kirk2022improving}. 

Finally, while we have explored varying the structure of the templates, the number of examples within a template, etc., in this paper we use only a single template for each type of query (goal or action) because the purpose of the paper is to compare the contributions of various knowledge sources (such as the LLM). Using the same template across all goals (and a similar one across all actions), for all conditions, limits sources of variation in results that might derive from more sophisticated or tuned template selection and obscure the contribution analysis.

To construct the prompt below, the agent instantiates the template by filling ``slots'' in the template with information from its current task context (\texttt{\Instantiate}, line 4). In this case, there are four slots (task, agent location, object of interest, object location) and it fills these slots with (respectively): \texttt{tidy-kitchen}, \texttt{kitchen}, \texttt{ceramic-plate}, and \texttt{on table}. The instantiated template is shown below. This example presents the specific text constructed by the agent that is sent to GPT-3. 

\begin{center}
\textbf{\emph{Agent-created prompt: Goal to tidy ceramic-plate?}}
\end{center}
\begin{verbatim}
(EXAMPLES)(TASK)Task name: store object.
Task context: I am in mailroom. Aware of package of office
supplies, package is in mailroom.
(RESULT)The goal is that the package is in the closet and
the closet is closed(END RESULT)
Steps:
1. Open closet
2. Pick up package of office supplies
3. Put package into closet
4. Close closet
(END TASK)
(TASK)Task name: deliver package. Task context: I am in mailroom.
Aware of package addressed to Gary, package is in mailroom.
(RESULT)The goal is that the package is in Gary's office
(END RESULT)
Steps:
1. Pick up package addressed to Gary
2. Go to Gary's office
3. Put package onto desk in Gary's office
(END TASK) 
(END EXAMPLES)
(TASK) Task name: tidy kitchen. Task context: I am in kitchen.
Aware of ceramic-plate on table.
(RESULT)
\end{verbatim}
\par
\par

Having instantiated the chosen template, the agent attempts to retrieve a number of unique responses (line 7). Algorithm~\ref{alg:unique_res} summarizes the retrieval process, which increases the temperature sent to the LLM with the prompt in successive presentations in order to attempt to produce as many unique responses as possible up to the desired number (i.e., \texttt{num\_responses}, which for goals in this experiment is set to 3). 

Below, we list the potential goals generated from the prompt above with this procedure. The first listed response was obtained for the temperature=0 response. The agent sends prompt queries with temperature=0.9 to attempt to retrieve two unique responses (for a total of three unique responses). If duplicate responses result from temperature=0.9 and there are still not three total unique responses, the agent will send more prompt queries with temperature=1.0, with the max total attempts set to 10. 

The purpose of these multiple queries with the same text is to attempt to retrieve three unique responses. The input temperature and resulting mean log probability of each response is listed following each response. GPT-3, with the logprobs=1 setting, reports the log probability of each token generated in the response. The mean log probability of a response is calculated by calculating the average of the log probabilities for each token in the response.

\begin{center}
\textbf{\emph{GPT-3 responses: Potential goals to tidy ceramic-plate}}
\end{center}
\begin{verbatim}
The goal is that the ceramic-plate is in the cupboard and the
cupboard is closed (temp=0, prob=0.915)
The goal is that the ceramic-plate is in the dishwasher and the
dishwasher is turned on (temp=0.9, prob=0.901)
The goal is that the ceramic-plate is in the sink and the sink is
full of water (temp=0.9, prob=0.866)
\end{verbatim}

To use these retrieved responses in this specific condition (i.e., with human confirmation), the next step in the process is to evaluate these responses (Algorithm~\ref{alg:finchoice}).  The agent sorts potential goals by the mean log probability to order the responses sent to the user so the most typical (highest probability given the GPT-3 training corpus) is presented first. The agent then asks the human user if a retrieved goal is correct for the object of interest (i.e., in this example, the plate), continuing through the choices until one is accepted or all are rejected (lines 7-12).  The user-agent dialogue is displayed below. Note that we include the tag [LM] to mark responses that the agent presents to the user that were generated from the GPT-3 language model.

\begin{center}
\textbf{\emph{User dialogue: Evaluating potential goals for ceramic-plate}}
\end{center}
\nopagebreak[4]
\begin{verbatim}
Agent: [LM] For a ceramic-plate on the table is the goal is that
the ceramic-plate is in the cupboard and the cupboard is closed?
Instructor: no.
Agent: [LM] For a ceramic-plate on the table is the goal is that
the ceramic-plate is in the dishwasher and the dishwasher is 
turned on?
Instructor: no.
Agent: [LM] For a ceramic-plate on the table is the goal is that
the ceramic-plate is in the sink and the sink is full of water?
Instructor: no.
Agent: What is the next goal or subtask of clear?
Instructor: If the object is a ceramic-plate then the goal is that
the object is in the dishwasher and the dishwasher is closed.
\end{verbatim}

In this case, the user rejects all three unique responses. The user rejects the first response because the user wants the agent to clean all plates that could be dirty. Because of the limited embodiment of the agent it cannot detect if a plate is clean or dirty.
The user rejects the second response because the desired outcome is to load all the objects in the dishwasher rather than to run the dishwasher after loading a single item. 
The third response is rejected because filling the sink is outside of the capabilities of the agent (to parse) and the embodiment (to manipulate faucets). Because all three responses retrieved from the LLM are rejected in this case, the agent asks the human to provide a goal description (line 14 of Algorithm~\ref{alg:finchoice}).

\subsection{Learning actions to achieve a goal}

After the agent has learned a goal for an object, it must learn what actions to take to achieve that goal (lines 15-34 of Algorithm~\ref{alg:main}). It has three options: search, ask the LLM, and ask the human. It attempts to associate actions with the goal in this order.

The agent uses a domain-general, iterative deepening algorithm for search, applying its primitives (\texttt{pick-up}, \texttt{put-down}, \texttt{open/close-door})\footnote{The robot also plans moves, such as moving to the table to grasp the plate, but these steps are implemented at run-time, at a level below the robotic primitives for search/planning. So while \texttt{pick-up} will require moving and grasping actions when executed, it is treated as a primitive action in the iterative deepening search.} to an internal representation of the state and then evaluating if the goal conditions are met. In this example, the goal provided by the human requires opening and closing the dishwasher, as well as picking up the ceramic plate and putting it into the dishwasher (4 steps). Therefore the agent fails to find a solution using a search depth limited to 2 steps. Note that for the search depth=4 condition, the agent would find a solution with search alone and no further steps would be needed for learning what actions satisfy the goal and then executing them (lines 19-20).

In response to the search failure, the agent attempts to retrieve potential actions from the LLM (line 24). Algorithm~\ref{alg:actions} is similar to the approach for goals (Algorithm~\ref{alg:goals}) but has two key differences. First, it attempts to learn only one action step at a time, even though multiple actions are (usually) necessary to achieve a goal. When the agent has access to search, it may only need to learn a single action for search to then be successful after the action is performed, so the algorithm is designed to elicit responses for single actions, execute them, and then search again. Second, in attempting to retrieve a single action, it first performs an LLM retrieval for a single word (line 11), assesses the resulting words, and then does another retrieval for the full action descriptions (line 21). In what follows, we walk through these steps individually.

Template selection and instantiation (lines 2-4) are similar to the process for goals. In this case, it chooses the action-eliciting template. The instantiated prompt includes \texttt{(RESULT)} text \texttt{(END RESULT)} (indicating what should be the ``result'' of performing these the steps). The agent instantiates this text with the goal statement it obtained above. The prompt resulting from instantiation is below.

\begin{center}
\textbf{\emph{Agent-created prompt: What actions can satisfy ceramic-plate goal?}}
\end{center}
\begin{verbatim}
(EXAMPLES)(TASK)Task name: store object.
Task context: I am in mailroom. Aware of package of office
supplies, package is in mailroom.
(RESULT)The goal is that the package is in the closet and
the closet is closed(END RESULT)
Steps:
1. Open closet
2. Pick up package of office supplies
3. Put package into closet
4. Close closet
(END TASK)
(TASK)Task name: deliver package.
Task context: I am in mailroom. Aware of package addressed
to Gary, package is in mailroom.
(RESULT)The goal is that the package is in Gary's office
(END RESULT)
Steps:
1. Pick up package addressed to Gary
2. Go to Gary's office
3. Put package onto desk in Gary's office
(END TASK)
(END EXAMPLES)
(TASK) Task name: tidy kitchen. Task context: I am in kitchen.
Aware of ceramic-plate on table.
(RESULT)If the object is a ceramic-plate then the goal is that
the object is in the dishwasher and the dishwasher is closed
(END RESULT)
Steps:
1.
\end{verbatim}

We now describe the two-step retrieval process for generating each action step (Algorithm~\ref{alg:actions}, lines 11-23). The agent sends the prompt as above and receives the top-five single word responses (using the logprobs=5 feature of the API as previously outlined). Although not guaranteed, in our experimentation, the first word generated by GPT-3 in response to this prompt template is usually a verb (action). Some of these actions are known to the agent (e.g., ``open'') and some are not (e.g., ``take''). The agent biases further generation of an action by preferring action words it knows using different thresholds for known and unknown words (lines 6-7). These thresholds were selected after some simple experimentation with a few objects, and were not tuned to the objects in the experimental data set.

The one-word responses to the prompt above are shown below. In this case, the top two choices are known to the agent and thus above the known-word threshold (i.e., $9\%$).

\begin{center}
\textbf{\emph{GPT-3 responses}: Possible actions to tidy a ceramic-plate}
\end{center}
\begin{verbatim}
Open (temp=0, prob = 0.549)
Pick (temp=0, prob = 0.206)
Check (temp=0, prob = 0.067)
Go (temp=0, prob = 0.065)
If (temp=0, prob = 0.027)
\end{verbatim}

For each word/token retained, the agent formulates a new prompt exactly like the previous one other than it also inserts the chosen starting action word. In the examples below, the prompt from above is now extended first with:

\begin{verbatim}
Steps: 1. Open
\end{verbatim}
and subsequently
\begin{verbatim}
Steps: 1. Pick 
\end{verbatim}

As for goal-eliciting prompts, the agent first sends a temperature=0 query to the LLM and then queries to retrieve temperature=0.9 responses (for a total of two unique responses). For actions, completion responses are often duplicates (even for high temperature). For both ``open'' and ``pick'' only one unique response is retrieved per starting word as detailed below.

\begin{center}
\textbf{\emph{Agent-created action completion prompt with ``open''}}
\end{center}
\begin{verbatim}
(EXAMPLES)(TASK)Task name: store object.
Task context: I am in mailroom. Aware of package of office
supplies, package is in mailroom.
(RESULT)The goal is that the package is in the closet and
the closet is closed(END RESULT)
Steps:
1. Open closet
2. Pick up package of office supplies
3. Put package into closet
4. Close closet
(END TASK)
(TASK)Task name: deliver package.
Task context: I am in mailroom. Aware of package addressed
to Gary, package is in mailroom.
(RESULT)The goal is that the package is in Gary's office
(END RESULT)
Steps:
1. Pick up package addressed to Gary
2. Go to Gary's office
3. Put package onto desk in Gary's office
(END TASK)
(END EXAMPLES)
(TASK) Task name: tidy kitchen.
Task context: I am in kitchen. Aware of ceramic-plate on table.
(RESULT)If the object is a ceramic-plate then the goal is that
the object is in the dishwasher and the dishwasher is closed.
(END RESULT)
Steps:
1. Open
\end{verbatim}
\begin{center}
\textbf{\emph{GPT-3 response to ``open'' prompt}}
\end{center}
\begin{verbatim}
dishwasher
\end{verbatim}

\begin{center}
\textbf{\emph{Agent-created action completion prompt with ``pick''}}
\end{center}
\begin{verbatim}
(EXAMPLES)(TASK)Task name: store object.
Task context: I am in mailroom. Aware of package of office
supplies, package is in mailroom.
(RESULT)The goal is that the package is in the closet and
the closet is closed(END RESULT)
Steps:
1. Open closet
2. Pick up package of office supplies
3. Put package into closet
4. Close closet
(END TASK)
(TASK)Task name: deliver package.
Task context: I am in mailroom. Aware of package addressed
to Gary, package is in mailroom.
(RESULT)The goal is that the package is in Gary's office
(END RESULT)
Steps:
1. Pick up package addressed to Gary
2. Go to Gary's office
3. Put package onto desk in Gary's office
(END TASK)
(END EXAMPLES)
(TASK) Task name: tidy kitchen.
Task context: I am in kitchen. Aware of ceramic-plate on table.
(RESULT)If the object is a ceramic-plate then the goal is that
the object is in the dishwasher and the dishwasher is closed.
(END RESULT)
Steps:
1. Pick
\end{verbatim}
\begin{center}
\textbf{\emph{GPT-3 response to ``pick'' prompt}}
\end{center}
\begin{verbatim}
up ceramic-plate from table
\end{verbatim}

The set of resulting retrieved actions (and mean log probability) for the first action step are:
\begin{verbatim}
Open dishwasher (prob=0.999)
Pick up ceramic-plate from table. (prob=0.944)
\end{verbatim}

As it did for goal responses, the agent now finalizes the response to use (Algorithm~\ref{alg:finchoice}). As before, the agent sorts the action statements by the magnitude of the mean log probability to determine the order of responses sent to the human instructor for confirmation. Importantly, in this environment, the agent has a single gripper, meaning it must first open the dishwasher before it picks up the ceramic plate and there is clear choice among the resulting options (i.e., that would not be true for a two-armed robot). The dialogue for the user-agent interaction is below.

\begin{verbatim}
Agent: [LM] For the ceramic-plate should I `Open dishwasher'?
Instructor: yes.
\end{verbatim}

After this action is confirmed by the human, the agent executes this instruction and opens the dishwasher (line 20 of Algorithm~\ref{alg:main}). The goal is not yet satisfied (line 15), and the agent now attempts again to search for the goal, now that it is (presumably) closer to the goal. However, again, the goal is more than 2 steps away and the search fails, causing the agent to construct another prompt to retrieve the next step to take from the LLM. The resulting process is the same as before, except that the prompts that the agent creates now include the action steps already retrieved and used by the agent. (The inclusion of prior action steps is part of the prompt instantiation process and not detailed in Algorithm~\ref{alg:actions}.)

\begin{center}
\textbf{\emph{Prompt for initial term after ``open dishwasher''}}
\end{center}
\begin{verbatim}
(EXAMPLES)(TASK)Task name: store object.
Task context: I am in mailroom. Aware of package of office
supplies, package is in mailroom.
(RESULT)The goal is that the package is in the closet and
the closet is closed(END RESULT)
Steps:
1. Open closet
2. Pick up package of office supplies
3. Put package into closet
4. Close closet
(END TASK)
(TASK)Task name: deliver package.
Task context: I am in mailroom. Aware of package addressed
to Gary, package is in mailroom.
(RESULT)The goal is that the package is in Gary's office
(END RESULT)
Steps:
1. Pick up package addressed to Gary
2. Go to Gary's office
3. Put package onto desk in Gary's office
(END TASK)
(END EXAMPLES)
(TASK) Task name: tidy kitchen.
Task context: I am in kitchen. Aware of ceramic-plate on table.
(RESULT)If the object is a ceramic-plate then the goal is that
the object is in the dishwasher and the dishwasher is closed.
(END RESULT)
Steps:
1. Open dishwasher
2.
\end{verbatim}

For this prompt, only a single response was generated that was above the thresholds as described previously: ``pick.''
\begin{center}
\textbf{\emph{GPT-3 response}}
\end{center}
\begin{verbatim}
Pick (temp=0, prob = 0.907)
\end{verbatim}

As above, the agent now creates a new prompt with ``pick'' appended as the first word of the second step in the prompt:

\begin{center}
\textbf{\emph{Prompt for action completion with ``pick''}}
\end{center}
\begin{verbatim}
(EXAMPLES)(TASK)Task name: store object.
Task context: I am in mailroom. Aware of package of office
supplies, package is in mailroom.
(RESULT)The goal is that the package is in the closet and
the closet is closed(END RESULT)
Steps:
1. Open closet
2. Pick up package of office supplies
3. Put package into closet
4. Close closet
(END TASK)
(TASK)Task name: deliver package.
Task context: I am in mailroom. Aware of package addressed
to Gary, package is in mailroom.
(RESULT)The goal is that the package is in Gary's office
(END RESULT)
Steps:
1. Pick up package addressed to Gary
2. Go to Gary's office
3. Put package onto desk in Gary's office
(END TASK)
(END EXAMPLES)
(TASK) Task name: tidy kitchen.
Task context: I am in kitchen. Aware of ceramic-plate on table.
(RESULT)If the object is a ceramic-plate then the goal is that
the object is in the dishwasher and the dishwasher is closed.
(END RESULT)
Steps:
1. Open dishwasher
2. Pick
\end{verbatim}
\begin{center}
\textbf{\emph{GPT-3 response}}
\end{center}
\begin{verbatim}
 up ceramic-plate
\end{verbatim}
 
The retrieved action (and mean log probability) for the second step (including ``pick'') is:
\begin{verbatim}
Pick up ceramic-plate (prob=0.99)
\end{verbatim}

All other responses were duplicates. The same as for the goals, if the user rejects this response, they will be asked to provide a description of the action to take. In this case the action is correct:

\begin{verbatim}
Agent: [LM] For the ceramic-plate should I 
            `Pick up ceramic-plate'?
Instructor: yes.
\end{verbatim}

After this action is confirmed by the user, the agent executes this instruction and picks up the ceramic-plate. The agent then attempts again to search for the goal. However, the goal is now only two steps away (\texttt{put-down ceramic-plate in dishwasher}, \texttt{close dishwasher-door}) and the iterative deepening search succeeds in finding the next actions to perform to achieve the goal, placing both of these actions on the \texttt{action\_list} (line 17 of Algorithm~\ref{alg:main}). The agent then executes the actions on the list. It first puts the ceramic-plate in the dishwasher and then it closes the dishwasher. This satisfies the goal for \texttt{ceramic-plate} and the agent moves to another object in Algorithm~\ref{alg:main}. 

The agent has now determined, through a process of search, constructing prompts to the LLM, and interaction with an instructor, what it should do when encountering a ceramic plate on the table when its task is to tidy a kitchen. At each step, as results are verified, the agent constructs new knowledge structures (in the form of production rules and semantic memory structures) that describes the goal and process of clearing the ceramic-plate on the table. This fine-grained learning allows the agent to immediately execute the steps to tidy the plate in the future, without resorting to the LLM or the human. (See next section for more details on what is learned and how it is used.)

The agent then moves on to the next object on the table and repeats the process, learning how it should tidy other objects it observes in the kitchen.

\section{Representation/use of learned task knowledge}

A distinguishing feature of this approach is that \textit{during} the initial execution of a task, while consulting a human and potentially a LLM, the agent \textit{learns} new procedural and declarative knowledge so that in future, when performing similar tasks, it does not require any search or interaction with a human or the LLM. That is the essence of its one-shot learning. This learning is possible because the agent incrementally stores procedural and semantic knowledge during its initial execution of the task that is used during future task execution. New task knowledge is encoded as fine-grained units that are not simply a memorization of the goal/action for a particular object but are generalizations based on the specific details of what was required to perform the task. These fine-grained units transfers to new, similar tasks as well as the identical one.

Declarative knowledge, stored in Soar's semantic memory \cite{laird_soar_2012_custom}, includes a representation of the goal (e.g., the conditions that the ceramic-plate is in the dishwasher and the dishwasher is closed) and the primitive actions used during the task (e.g. `open,' `pick-up,' `put-down,' and `close'). This declarative specification of goal and action is learned as the agent receives the goals (from the LLM or the user) and actions (from search, the LLM, or the user). The declarative task-structure was defined in previous research on task learning \cite{mininger_expanding_2021}. This research uses this identical declarative learning paradigm and task structure, although it was originally defined for learning only from direct user interaction.

Procedural knowledge, represented as rules in Soar, includes knowledge for selecting the next goal or subtask to perform for a given task, and policy knowledge to propose relevant actions and select which action to perform in a given task situation. Procedural knowledge is learned while executing the task actions and during a retrospective analysis of how the executed actions achieve the goal. This retrospective analysis consists of recreating the initial situation, using episodic memory to replay the execution, and then doing a causal-dependency analysis to determine which aspects of the state and goal were necessary for selecting each action on the path to the goal. The dependency analysis is a form of online explanation-based generalization \cite{dejong_explanation-based_1986,mitchell_explanation-based_1986}implemented in Soar's procedural learning mechanism, chunking \cite{laird_soar_2012_custom}. The dependency analysis ensures that only those aspects of the state, actions, and goal that were necessary for achieving the goal are included in learned rules. 

Because of the generalization made possible by the dependency analysis, learned knowledge transfers to similar situations. In some cases, transfer results in future search needing fewer search steps (i.e., the agent can recognize earlier in the search whether a search path leads to the goal or not).  Below, we illustrate some of these cases, drawn from procedural knowledge learned during learning to tidy, or clear, the ceramic-plate (i.e., continuing the example used in the prior section). We describe the procedural knowledge learned for the open action; similar knowledge is learned for other actions as well.

The agent first learns to select an appropriate goal when the task is to clear and the category of the object of interest is a ceramic-plate.\footnote{The basic processing unit of Soar is an operator. An operator's processing is performed by independent rules that propose, evaluate, and execute the operator. For simplicity, we elide some of these distinctions. In a technical sense, the agent is learning new operators that perform various tasks, such as selecting and retrieving goals and executing external actions.} A simplified version of the rule with the conditions listed first (after ``If") and the result of the rule matching (after ``Then") is depicted below. Subsequent examples of rules follow the same pattern. \texttt{O1} is an \textit{identifier} in Soar and is used to bind together multiple properties and relations for a given object, in this case an observed object that has category ceramic-plate. The current\_task of clear has the object \texttt{O1} as the primary argument of the proposed task. The proposed next-goal \texttt{clear1goal24} refers to a unique identifier that the agent generated for the goal it learned for clearing the ceramic-plate (that it is in the dishwasher and the dishwasher is closed).

\begin{center}
\textbf{\emph{Procedural Rule for ceramic-plate goal selection}}
\end{center}
\begin{verbatim}
If current_task = clear(O1) and detected object O1
AND category(O1) = ceramic-plate
Then:
propose select-next-goal clear1goal24
\end{verbatim}

After the agent acquires the goal for clearing the ceramic-plate on the table, and retrieves the first action to execute (from the LLM confirmed by the user), the agent learns to propose ``open the dishwasher." The conditions for this proposal are that the task is clear, the dishwasher is closed, and the agent is not holding an object. The knowledge is relevant for any goals for clearing the table that involve the dishwasher and will apply for other objects. A simplified representation of the rule learned is given below. Similar rules, also shown, are learned for the other actions, `pick up,' `put down,' and `close.' The rules for pick up and put down will apply to any object with an affordance where the robot can manipulate (``grab") them.

\newpage

\begin{center}
\textbf{\emph{Procedural Rule for Open proposal}}
\end{center}
\begin{verbatim}
If current_task = clear and detected object O1
AND category(O1) = dishwasher and property(O1) = closed
AND holding_object = false
Then:
propose action Open(O1)
\end{verbatim}

\begin{center}
\textbf{\emph{Procedural Rule for Close proposal}}
\end{center}
\begin{verbatim}
If current_task = clear and detected object O1
AND category(O1) = dishwasher and property(O1) = open
AND holding_object = false
Then:
propose action Close(O1)
\end{verbatim}

\begin{center}
\textbf{\emph{Procedural Rule for Pick up proposal}}
\end{center}
\begin{verbatim}
If current_task = clear(O1) and detected object O1
AND holding_object = false
AND property(O1) = not_grabbed and affordance(O1) = grabbable
Then:
propose action Pick-up(O1)
\end{verbatim}

\begin{center}
\textbf{\emph{Procedural Rule for Put-down proposal}}
\end{center}
\begin{verbatim}
If current_task = clear(O1) and detected objects O1,O2
AND property(O1) = grabbed
AND property(O2) = open and affordance(O2) = receptacle 
                        and category(O2) = dishwasher
Then:
propose action Put-down(O1,O2)
\end{verbatim}

When the agent attempts to execute open dishwasher, if the action cannot be immediately executed, which is the case if the agent is not directly next to the dishwasher, it performs internal search over primitive actions. The agent learns to select (meaning to execute this proposed action) the approach action if it is attempting to open an object (that has the property of ``openable") and the object is not ``reachable" (not close enough for the agent to interact with the object). This generalized knowledge applies to other objects, such as the pantry, that must be opened as well as the dishwasher, so that in the future no additional learning is necessary. The resulting (simplified) rule is:

\begin{center}
\textbf{\emph{Procedural Rule for Approach selection}}
\end{center}
\begin{verbatim}
If current_action = Open(O1) and proposed_action = Approach(O1)
AND property(O1) = not_reachable
Then:
select action Approach(O1)
\end{verbatim}

After the retrospective analysis of the actions that were used for the ceramic-plate, the agent learns to select the open action for dishwasher (which is closed) if it is performing the clear task on an object that is not currently grabbed, and where the current goal is that the object is in the dishwasher and the dishwasher is closed. 
 This rule includes more tests (is more specific) than the previous rules, but still generalizes the particular object being cleared. This rule, shown below, will apply for clearing other objects into the dishwasher. Further generalization is possible (future work) so that the rule will transfer to other ``closeable" objects (i.e., not just the dishwasher).

\begin{center}
\textbf{\emph{Procedural Rule for Open selection}}
\end{center}
\begin{verbatim}
If current_task = clear(O1) and proposed_action = Open(O1)
AND detected objects O1,O2
AND category(O2) = dishwasher and property(O2) = closed
AND property(O1) = not_grabbed
AND current_goals = {O1 in O2, property(O2) = closed}
Then:
select action Open(O1)
\end{verbatim}

Similar proposal and selection rules are learned for ``pick up," ``put-down," and ``close" during action execution and retrospective analysis. These rules define a state-based policy rather than a rigid procedure specifying a list of actions to perform. Thus, in future scenarios, not only will the agent have the necessary knowledge to perform clear on the ceramic-plate (without any further search or queries to the LLM or user), it also has knowledge that can apply (transfer) for handling similar objects and situations.

For example, when the agent encounters the metal-fork on the table and learns that the goal is that the metal-fork should be in the dishwasher with the dishwasher closed, it does not need to learn any new actions or policy knowledge to clear the metal-fork. It immediately proposes and performs the actions to achieve the goal, again without using search or querying the user or LLM. This ability to learn fine-grained task knowledge and apply it to future performance, as well as generalize knowledge for similar situations, is a key difference between our approach and the approaches that learn new tasks by memorization of instruction and/or exploration in the domain.

\end{document}